\documentclass[letterpaper, 10 pt, conference]{ieeeconf}  

\IEEEoverridecommandlockouts                              

\usepackage{mwe}
\usepackage{algorithm2e}
\usepackage{xcolor}
\usepackage{url}
\usepackage{hyperref}
\usepackage{breakurl}
\usepackage{verbatim}
\usepackage{bera}
\usepackage{listings}

\colorlet{punct}{red!60!black}
\definecolor{background}{HTML}{EEEEEE}
\definecolor{delim}{RGB}{20,105,176}
\definecolor{mary}{RGB}{255,0,0}
\colorlet{numb}{magenta!60!black}

\lstdefinelanguage{json}{
    basicstyle=\normalfont\ttfamily,
    numbers=left,
    numberstyle=\scriptsize,
    stepnumber=1,
    numbersep=8pt,
    showstringspaces=false,
    breaklines=true,
    frame=lines,
    backgroundcolor=\color{background},
    literate=
     *{0}{{{\color{numb}0}}}{1}
      {1}{{{\color{numb}1}}}{1}
      {2}{{{\color{numb}2}}}{1}
      {3}{{{\color{numb}3}}}{1}
      {4}{{{\color{numb}4}}}{1}
      {5}{{{\color{numb}5}}}{1}
      {6}{{{\color{numb}6}}}{1}
      {7}{{{\color{numb}7}}}{1}
      {8}{{{\color{numb}8}}}{1}
      {9}{{{\color{numb}9}}}{1}
      {:}{{{\color{punct}{:}}}}{1}
      {,}{{{\color{punct}{,}}}}{1}
      {\{}{{{\color{delim}{\{}}}}{1}
      {\}}{{{\color{delim}{\}}}}}{1}
      {[}{{{\color{delim}{[}}}}{1}
      {]}{{{\color{delim}{]}}}}{1},
}

\newcommand{\sa}{droidlet agent}
\newcommand{\pf}{droidlet platform}
\newcommand{\pgf}[1]{\newline \noindent {\bf #1}}
\title{\LARGE \bf
droidlet: modular, heterogenous, multi-modal agents
}

\author{Anurag Pratik$^{1}$ Soumith Chintala$^{1}$ Kavya Srinet$^{1}$ Dhiraj Gandhi$^{1}$
Rebecca Qian$^{1}$ Yuxuan Sun$^{1}$ Ryan Drew$^{*}$ \\ Sara Elkafrawy$^{*}$ Anoushka Tiwari$^{*}$ Tucker Hart$^{*}$ Mary Williamson$^{1}$ Abhinav Gupta$^{1,2}$ Arthur Szlam$^{1}$

\thanks{*Work done while at Facebook AI Research}
\thanks{$^{1}$Facebook AI Research}%
\thanks{$^{2}$Carnegie Mellon University}%
}

\begin{document}

\maketitle
\thispagestyle{empty}
\pagestyle{empty}

\begin{abstract}
 In recent years, there have been significant advances in building end-to-end Machine Learning (ML) systems that learn at scale. But most of these systems are: (a) isolated (perception, speech, or language only); (b) trained on static datasets. 
 On the other hand, in the field of robotics,  large-scale learning has always been difficult. Supervision is hard to gather and real world physical interactions are expensive. 
 
In this work we introduce and open-source droidlet, a modular, heterogeneous agent architecture and platform. It allows us to exploit both large-scale static datasets in perception and language and sophisticated heuristics often used in robotics; and provides tools for interactive annotation.  Furthermore, it brings together perception, language and action onto one platform, providing a path towards agents that learn from the richness of real world interactions. 
 \end{abstract}

\section{Introduction}
A long-standing goal of AI research has been to build learning systems that can continuously improve themselves through interaction with their environment and through interactions with humans.  However, agents that can flexibly learn without human engineers carefully teeing up their training data still exist only in science fiction.

One approach that has shown promise is to build virtual agents in simulations.  
Simulations allow researchers to explore self-supervised and self-directed learning agents that have access to large data, and so build on recent advances in ML that have demonstrated great success in the large-data setting. However, despite the expanding sophistication of simulated environments,
there is still the danger that agents only learn what humans encode into the simulation.

An alternative is to use robots and the richness of real world interactions. However, performing large-scale interactive learning on 
real-world robots is difficult. Kinesthetic/Teleop supervision is expensive, the real world is slow (compared to simulations), and robots break and get stuck. 

\begin{figure}[t!]
\includegraphics[width=\columnwidth]{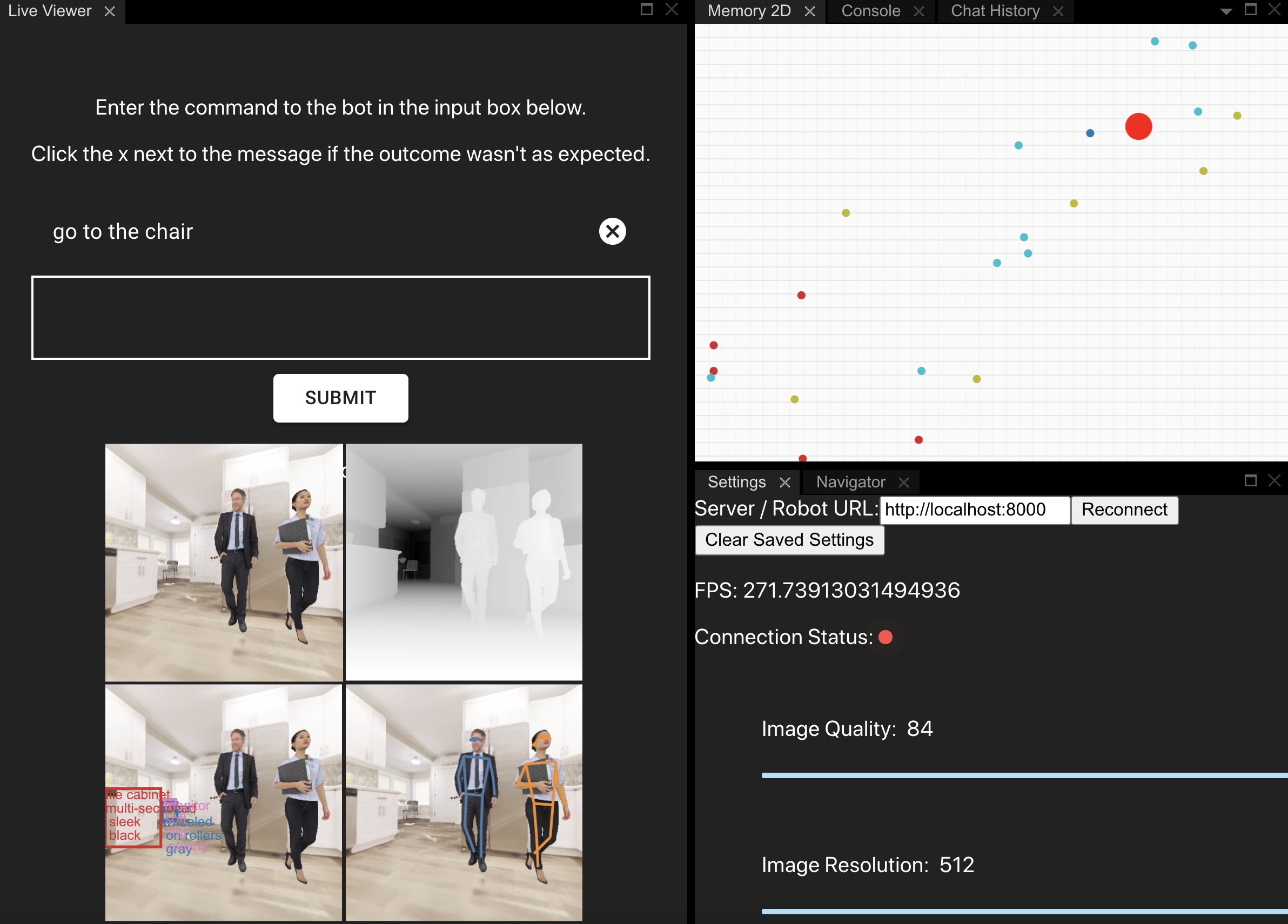}
\caption{A screenshot of the dashboard (\ref{sec:dashboard}). On the top left is a text box to send commands to the bot. The 4 sub-windows on the left are the image stream (center-left), depth map (center-right), detector output (bottom-left) and human pose estimation (bottom-right). On the right side is the Memory pane which has a 2D view of all objects in memory, and other panes available for debugging and interacting with the agent. \label{fig:dashboard}}
\end{figure}

Our view is that we can resolve this dilemma through agents with modular designs and heterogenous supervision.  Rather than thinking of ``the'' agent as a monolith, we consider a family of agents, made up of a collection of modules \cite{andreas2016neural, devin2017learning, dalmia2019enforcing} some of which may be heuristic, and others learned.   
We believe this makes both scaling tractable, and improves sample efficiency where scale is not possible:
\pgf{Scaling} A carefully designed agent architecture and platform allows {\it components} to train on large data when large data is available for that component; and to use sophisticated heuristics when good programmers can provide them.  All of the components need not get all the experience; and static data can be used as appropriate.
\pgf{Sample Efficiency and Self-Supervision}
We embrace and exploit the richness and multi-modality of real-world interactions, combining perception, action and language onto one common platform.  We can leverage the advances in ML models of each of these modalities to build useful high-level representations. Self-supervision is then about understanding the relationships between the modalities rather than focusing on the representations themselves.

In short, where large data is available for a module, use it; and when it is not, leverage the synergies of the modules.
We believe this approach is complementary to simulating at scale!








\section{droidlet}
In this work 
we introduce and open-source the \pf{} and \sa{} (an agent instantiation that forms the core of the platform).  These allow a robot to meaningfully interact with the world and people in the world using current technology, but also provide a substrate for exploring new learning methods.  The platform and agent provide
\begin{itemize}
\item Primitives for language
\item APIs linking memory, action, perception, and language
\item Suite of perceptual models
\item Suite of data annotation tools.
\end{itemize}
The \sa{} is designed to be modular,  allowing swapping out components (which can be learned or heuristic) and extending the agent by adding new components.  Our goal is to build an ecosystem of language, action, memory, and perceptual models and interfaces.



In the rest of this document we give a tour of the platform.  In Section \ref{sec:agent} we describe the agent architecture in terms of its modules and how they interact.  Then in Section \ref{sec:platform} we discuss other components of the platform.  These include user interfaces for interacting with agents, annotating training data, and inspecting agent state; and a model zoo with various pre-trained ML models.  The code and data are located at \url{https://github.com/facebookresearch/droidlet}.

\section{Agent Design}
\label{sec:agent}
\begin{figure}[h!]
\includegraphics[width=\columnwidth]{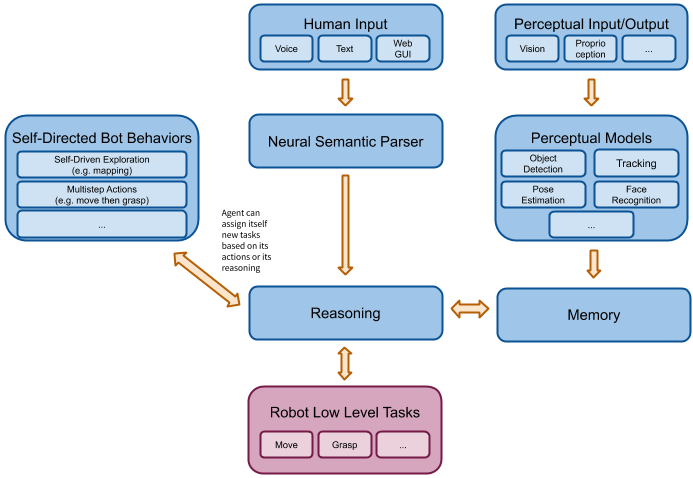}
\caption{Major components of the \pf and relationships between each. Modularity allows learned or heuristic modules to be swapped in or out for each. In addition, the platform is easily extended as described by the unnamed modules in each area.  \label{fig:agent_arch}}
\end{figure}

Our basic agent design is made up of a memory system, a task queue, a perceptual API, and a controller (a dialogue queue).  The core event loop of the Agent is shown in Algorithm \ref{alg:core_loop} and is diagrammed in Figure \ref{fig:agent_arch}.  Note that the components can be used separately from the main agent; or can be replaced or combined as a user sees fit; we discuss this more in \ref{sec:platform}.    The agent can be instantiated in any setting where the abstract functions defined in the perceptual and low-level action APIs can be implemented.   We release examples on a Locobot, using PyRobot \cite{pyrobot} as the low level controller in the real-world and in Habitat \cite{habitat19iccv}, and in Minecraft using \cite{craftassist} as the low-level controller. The associated code can be found at \url{https://github.com/facebookresearch/droidlet}.

In the next sections we discuss the components of the system individually, and then review how they interact.  At a high level, the principal design goal was to make the agent modular in such a way that the modules could be learned via ML or scripted; and so the modules could be used separately, or merged.
\begin{algorithm}
\SetAlgoLined
 \While{True}{
  run fast perceptual modules\;
  update memory\;
  step controller:\;
  \Indp 
  check for incoming commands \;   
  step highest priority object on the Dialogue queue\;
  \Indm
  step highest priority task\;
 }
 \caption{Core agent event loop \label{alg:core_loop}}
\end{algorithm}

\subsection{Dialogue Queue}
The agent maintains a {\it Controller}, whose job it is to place tasks on the {\it Task Queue} (described in the next section).  In the Locobot and Craftassist example agents, the Controller is further broken down into {\it Dialogue Queue} and a {\it Dialogue Controller}.  The Dialogue Controller is responsible for putting {\it dialogue objects} into the Dialogue queue; when stepped, these objects interpret human utterances and place tasks in the task queue, or ask for clarifications etc. (and so put other dialogue objects on the dialogue queue).   If the dialogue queue and task queue are empty, the agent may engage in information seeking behavior either by asking questions or otherwise; mechanically this is done by placing an object mediating that behavior on the appropriate queue.  

The current dialogue objects themselves are heuristic scripts
.  For example, there is a dialogue object that describes the state of the agent's task queue to the agent's interlocutor, which simply makes a memory call and formats the output.  Another asks for clarification about a reference object. 
Researchers focusing on ML for dialogue could replace these dialogue objects with learned versions (or add others, or combine all into one neural model...).  Our hope is that this basic architecture will even allow users to teach new dialogue objects on the fly, although the current \sa{} does yet have this capability.  Nevertheless, this architecture does make it simple to mix scripted with ML-directed dialogue.

\begin{figure}
\includegraphics[width=\columnwidth]{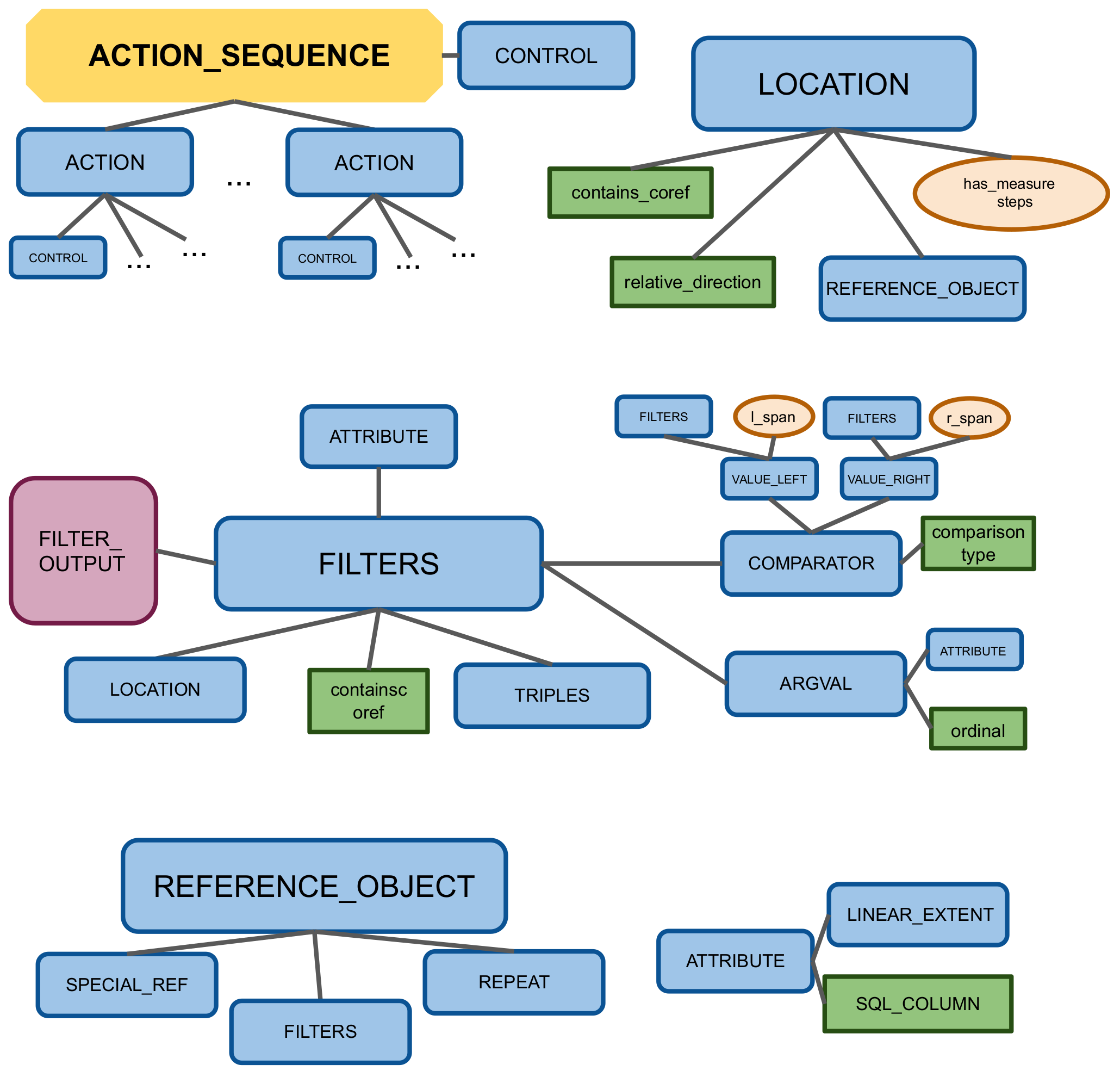}
\caption{A simplified diagram showing a part of the \sa{}'s DSL.  Rectangles with rounded corners are nodes with possible children, not always shown.  Rectangles with sharp corners are ``categorical'' leaves, with a fixed set of possible values (also not shown); and ovals are ``span'' leaf nodes that take values by copying from input text.  The complete DSL is described at \url{}  \label{fig:DSL}}
\end{figure}

The dialogue controller is mediated by a neural semantic parser (see \ref{sec:NSP} for more details).  It takes human utterances and translates them into partially-specified programs in a domain specific language (DSL).  These are then fully specified by the {\it interpreter} dialogue object.    A subset of the DSL is cartooned in Figure \ref{fig:DSL}; for a full description visit \url{https://bit.ly/3jMGhkn}.

\subsubsection{Interpreter}
The most involved dialogue object in the example \sa{}s is the interpreter.
This (currently heuristic, not learned) object uses the memory system as the interface to the perceptual system and thus to the "world".  Using information from memory, the interpreter chooses Tasks to put on the Task queue. 

The interpreter itself is modular, mirroring the structure of the agent's DSL.  The composable modules of the interpreter could be independently replaced with learned versions.

\subsection{Task Queue}
The agent's Task Queue stores Tasks, which are (mostly) self-contained lower-level world interactions.   Examples Tasks are moving to a specified coordinate in space, turning to face a particular direction, pointing, etc.   The execution of the \sa{}'s Tasks are mostly heuristic at this point, with the exception of grasping using the arm (see \ref{sec:grasp_model}).

As in other components of the system, Tasks and the Task Queue are designed to allow easy mixing of heuristic and learned modules.

\subsection{Memory}
The memory system stores and organizes information for the agent.  In addition to memory in the sense of ``recollection'' of past information, it is the interface between the perceptual system and the Controller. The high-level interface to the memory consists of {\it memory nodes}, and a subset of the DSL that formalizes possible queries over the memory nodes. 

\begin{figure}[h!]
\includegraphics[width=\columnwidth]{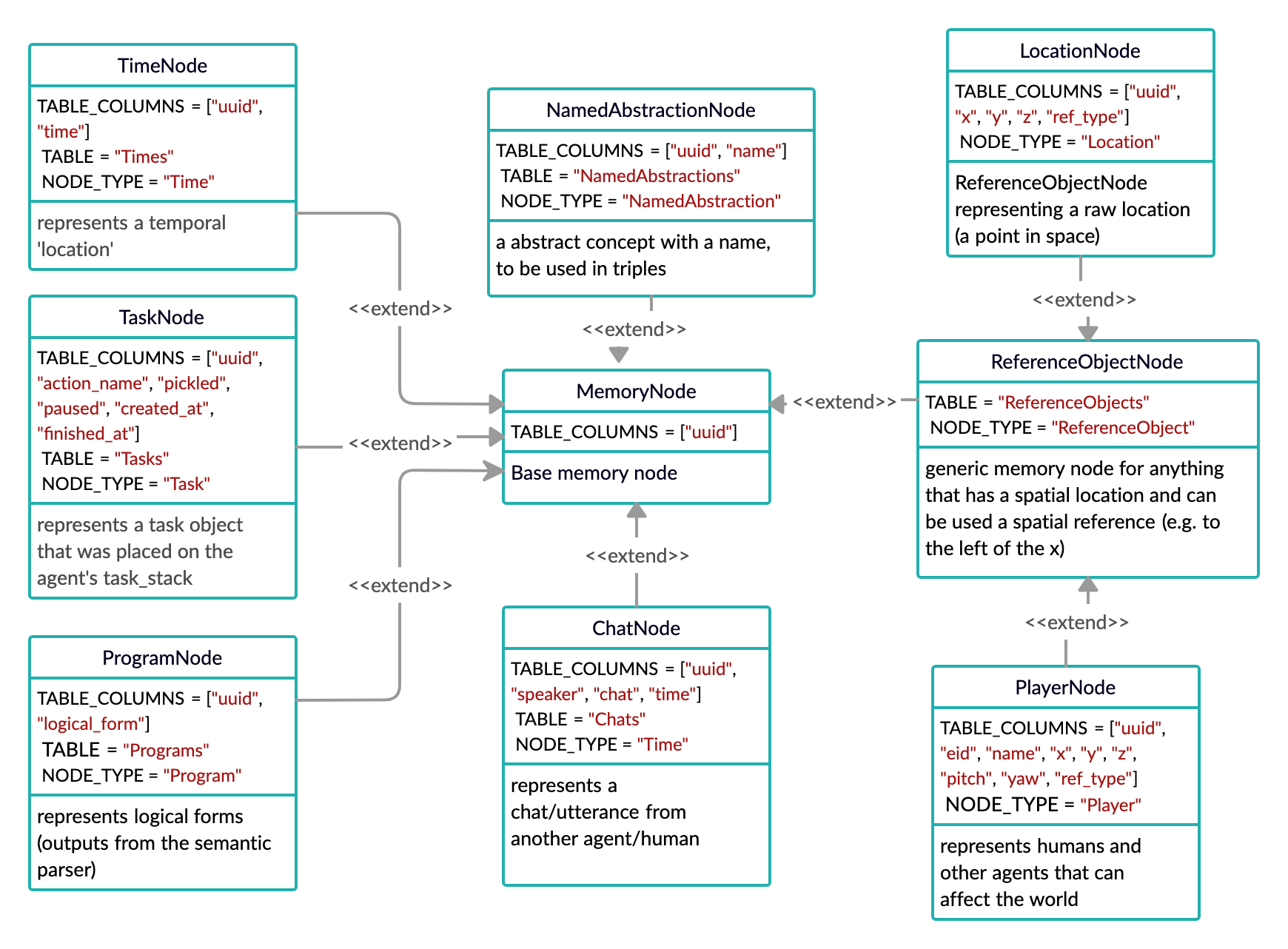}
\caption{Some memory nodes \label{fig:memnodes}}
\end{figure}

 There are memory nodes that represent reference objects (that is, things with a spatial location), temporal objects (things with a location in time), (the text of) utterances of the agent or interlocutors, programs the agent has run or is running, and Tasks on the Task Queue. 
 As the memory system also has support for  semantic triples (e.g. {\it $<$particular\_chair$>$, has\_colour, red}), there are memory nodes for more conceptual objects like
  sets of memory nodes,  and abstract entities (e.g. the class of chairs, as opposed to that $<$particular\_chair$>$). For each object there is also an archive type, for storing the state of a memory object at a moment in time; many of the perceptual memory nodes only store the most recently observed information.   Some examples are shown in Figure \ref{fig:memnodes}.

  All memory nodes are equipped with a unique identifier and time-stamps of creation time and last access.  While the ``backbone'' of the memory system is discrete in the sense that memory data is stored in SQL tables, addressable by the identifier, we can also store floating point vectors (e.g. locations, poses, features from ML models) as table data.  


\subsection{Perception Interface}
At a high level, any perceptual information to be used by the controller should be stored in memory, and accessed through memory nodes and the memory-query subset of the DSL.  This arrangement makes it easier to swap out perceptual modules and add new ones, and allows that agents with different underlying hardware and perceptual capabilities can use the other aspects of the agent.

Several of the perceptual modules included in the agent are primarily visual: object detection and instance segmentation, face detection and recognition, and pose estimation.   Their memory interfaces are also supported by a deduplication system. For example, for each object detection the agent makes, it is necessary to decide whether that detection is an entirely new object or a different view of an object already recorded in memory, to be updated.   In the Locobot instantiation, these are powered by learned models described in further detail in Section \ref{sec:vision_models}.  Others modules support the agent's proprioception, e.g. localization and self-pose.

In our PyRobot instantiation, some modules can use multiple ``modalities''; for example to decide where an interlocutor is attending.  This is currently mediated either by vision (via detecting the dot of a specially prepared laser pointer \ref{sec:vision_models} {\it or} through the dashboard (Section \ref{sec:dashboard}).  The important thing here is that from the perspective of the interface, the mechanism by which the perceptual model or heuristic obtains the data (or even if there is a single source of data) is irrelevant; all that matters is that data is put into memory in the correct format.


\subsection{Putting it Together}
We now step through the \sa{}'s processes as it handles a command from a person, who tells it to "go to the chair".  To begin with, we need to explain how the person told the agent something, as the current instantiation of hardware running the \sa{} does not have microphones, and we have not discussed audio perception\footnote{The agent design happily extends to adding audio perception or audio sensory hardware.}.  Instead, the message is passed through the dashboard app; see Section \ref{sec:dashboard}.  

Because the agent has received a message, it makes a more detailed perception check.  The semantic parser is then run on the message, outputting the logical form\footnote{After handling text spans; see \cite{caip2020}}
\begin{lstlisting}[language=json, numbers=none]
{"dialogue_type": "HUMAN_GIVE_COMMAND",
 "action_sequence": [
    {"action_type" : "MOVE",
     "location": {
      "reference_object":{
          "filters": {
            "has_tag": "chair" }}}}]}
\end{lstlisting}
Because the semantic parser (which doubles as the controller here) labeled the chat as being of type ``HUMAN\_GIVE\_COMMAND'',  the logical form is fed to an Interpreter dialogue object which is placed on the Dialogue Queue.  The Interpreter instantiates a memory search specified by the ``filters'' sub-dictionary of the logical form.   If the agent has at any point detected an object that it classified as a ``chair'' (or if some detected object has been tagged by a human as having the label ``chair'', either through dialogue or through the dashboard's labeling tools), it will be in the memory with the proper tag; and the interpreter will place a move Task onto the Task Queue with target coordinates given by the detected object's location\footnote{If there are more than one, the agent defaults to the one nearest the human's point/looking-at location (if that is in memory), and otherwise the one nearest the agent}.  If there is not one in memory, the Interpreter places a Clarification dialogue object on the Dialogue Queue.

\section{Platform}
\label{sec:platform}
This project builds towards more capable high-level robotic agents through modularity.  If one contributor is able to improve a piece of the system, it should improve all the robotic agents using that piece of the system.  The goal of the \pf{} is to make it so these improvements are practical (as opposed to theoretical).  We wish to remove friction in integrating new ML models and new capabilities (scripted or learned) into the system.  The \sa{} described above should be considered an example of how to build a system using the components; but we do not consider the \sa{} to be the only (or even the main) way the \pf{} should be used.

In the next sections, we discuss the ``agent'' from the perspective of the inter-operability of its components, and how these can plug-and-play with other projects.

\subsection{Dashboard}
\label{sec:dashboard}
A common need across many robotics systems is a tool for inspecting the systems internal state.  For ML pipelines, tools are needed for annotating data.  The {\it dashboard} serves these purposes.  It can be used for debugging and visualization, correcting agent errors on-the-fly, and rich data annotation, and as a operational interface with agents.  In Figure \ref{fig:dashboard} we show a screenshot of the dashboard, with some of the panes described in sections \ref{sec:remote_control} and \ref{sec:dashboard_diagnostics}.
As with the rest of the agent, we prioritize modularity, and it is easy for a user to add new panes and tools.
\subsubsection{Remote Control}
\label{sec:remote_control}
The first use of the dashboard is as a control interface.  A user can directly teleop the agent, or communicate with the agent via text chats or speech.
\subsubsection{Diagnostics and State Visualization}
\label{sec:dashboard_diagnostics}
The dashboard also can be used for debugging or otherwise viewing the agent's internal state.  There are panes
 showing the state of the agent's memory system: a 2-d spatial map with reference objects marked in the locations where the agent has recorded them, views of the agents Task and Dialogue queues, and recently parsed logical forms.  Other panes show the outputs of the agent's perceptual modules, for example instance segmentation or pose estimation.
\subsubsection{Annotation Tools}
\label{sec:dashboard_annotation}
The dashboard has several panes that can be used to annotate data for ML pipelines.
\pgf{Vision annotation}  Support for instance segmentation (and labeling properties, tags, and labels of the segments), pose-estimation, and face detection.  The tool also tracks objects across frames and allows annotating videos.  
\pgf{Program annotation} A blockly interface into the agent's DSL, with support for naming and describing programs in natural language.  This tool is inspired by 
\cite{vash2020, wang2017naturalizing}, and as in those cases, it becomes more powerful as more people use it, and contribute annotated programs.
\pgf{Semantic parser annotation} From \cite{caip2020};  simplifies annotating a parse of an utterance to a logical form, using the structure of the DSL.

Each of the annotation tools can be used either live, interacting with the robot; or offline on stored data, allowing efficient crowd-work tasks.  
Our long-term goal is to match all diagnostic panes with annotation panes, allowing annotation whenever inspection of the agent's state reveals an error.

\subsection{Model and Data Zoo}
Swapping out ML models (or heuristics!) that conform to the requisite APIs should be straightforward.  For example, if a researcher finds that the instance segmentation model packaged with the \sa{} does not work well in their lab (maybe because of the distribution of objects  or the camera hardware on their robot, etc.), it should be easy to replace the model with one they train themselves.   Furthermore, we want 
researchers and hobbyists to share their models and data.  To support this, the \pf{} has model and data zoos, allowing contributors to train and post new models (or heuristics), and contribute new training data.   The current models in the zoo (supporting the \sa{}) are:
\subsubsection{Neural Semantic Parser}
\label{sec:NSP}
The neural semantic parser inputs text and outputs logical forms in the agent's DSL.  The version in the \sa{} uses the same architecture as the BERT based model in \cite{caip2020}, but the data has been extended to cover the new additions to the DSL beyond the grammar in \cite{caip2020}.  In particular, the DSL has been extended to support grasping actions, more powerful memory searches, and more powerful conditions. See \url{} for the \sa{}'s complete DSL.

\subsubsection{Vision models}
\label{sec:vision_models}
 We provide an interface to make it easy to string together task-specific vision modules and a multiprocessing design to even run them at different frame rates.
\pgf{Instance Segmentation and Object Properties} - we provide a modified Mask R-CNN \cite{maskrcnn} with an additional head that predicts object properties, pre-trained on LVIS \cite{gupta2019lvis} and fine-tuned on our local dataset collected through our suite of annotation tools for 38 object classes and 298 object properties.\footnote{Data was collected using a robot built by Facebook, in a Facebook office. All persons depicted in the images have provided their permission to have their image captured and made available to the public.}
\pgf{Real-time 2D object tracking} to track each detection in-frame using a SORT \cite{rt-track} based algorithm.
\pgf{Human Pose Estimation} using a pre-trained COCO \cite{coco} keypoint detector.
\pgf{Face Recognition} using a combination of the face recognition module from the dlib python library and Multi-task Cascaded Convolutional Networks \cite{facenet}.
\pgf{Laser Pointer handler} built from a basic vision heuristic of color filtering and contour detection to help with proprioception, for example "look at where I am pointing".
\pgf{Memory Deduplication} - we provide a basic heuristic using convolutional features and object locations to predict whether a detected object has been seen before despite possible changes in position, orientation, etc.

\subsubsection{Grasping}
\label{sec:grasp_model}
To grasp objects we used the learning-based grasping algorithm deployed in PyRobot API \cite{pyrobot}. The grasping model is trained using the data collected from people's homes \cite{robotsinhome}. The model takes an RGB image from the camera as input and outputs a grasp in image space. The grasp prediction consists of a 2D pixel location and the gripper orientation. This predicted 2D location is then converted to a real-world 3D location using the known camera parameters and the depth image. 
\subsection{Extensibility}
In addition to swapping out components of the agent that already exist, each of the parts of the agent architecture have been designed designed to be extensible with new ML or heuristic components.  For example, in order for a researcher or hobbyist to add a new perceptual modality, on the agent side, they need only add the interface to the agent's memory.  
\section{Related work}
\textbf{Lifelong and Self-supervised Learning in Real World.} One of the long term goals of our platform is to facilitate building of agents that learn in real world. The key features are scaling in terms of size, richness, multi-modality, human interactions and time horizon over which the system learns. In recent years, self-supervised learning has picked up pace in several sub-fields of AI including computer vision~\cite{Doersch2015, Wang2015, MoCo, SimCLR}, natural language processing~\cite{BERT, word2vec}, and even robotics~\cite{Pinto2016Grasping, chelseapushing, pulkitpoking}. However, most of these systems still use static datasets. Our platform facilitates building agents that interact with their environments and collect their own data. Lifelong learning has been a topic of interest in several subfields of AI. This has led to some impressive systems in both NLP~\cite{NELL} and Vision~\cite{NEIL}. However, most of  these efforts still use internet data and not the real-world interactions. While recent attempts in robotics have attempted to move out of lab and use diverse data collected in homes~\cite{robotsinhome}, the data is still collected in controlled settings. Our work is partly inspired from ~\cite{blenderbot, craftassist} where the aim is to use real-world interactions with users and build learning systems that learn from these interactions. However, these systems still do not exploit the richness of real world interactions and the presence of multiple modalities. Our effort attempts to bring together vision, action and language into one framework for learning.

\textbf{Research Ecosystems in AI Fields.} In recent years, research in AI has benefited immensely from research ecosystems such as PyTorch~\cite{pytorch} and TensorFlow~\cite{tensorflow}. These research ecosystems reduce the entry barrier by providing modular architectures with reusable components~\cite{pytorchmodules, tfmodules}, good practices for building systems~\cite{pytorchtutorial, tftutorial} and in many cases trained models on public datasets~\cite{modelzoo}. Apart from core ML generic ecosystems, there have also been growing plethora of specialized research toolkits as well. Some successful examples include Detectron~\cite{Detectron} in Computer Vision, AllenNLP~\cite{Gardner2017AllenNLP} in NLP, PyRobot~\cite{pyrobot} in robotics but most of these platforms only aim to impact a single sub-field in AI. We believe as we build agents that live and interact in the real world, we need platforms that can exploit the richness of real world interactions by bringing together different sub-fields in AI under one common platform. While platforms such as MMF~\cite{mmf} aim to bring together vision and language research, it still lacks the action and real-world physical interactions. An agent that interacts with physical world including humans and objects would need perception, action, ability to interact via language and memory to capture context. The \pf{} facilitates building agents that learn from the richness (and structure) of our physical and social world.

\textbf{Robotics Platforms.} The robotics community has embraced and preferred modular layered design~\cite{brooks} and reusable modules~\cite{thinsoftware} from early days. These principles led to development of one of the largest robotics ecosystem called ROS~\cite{quigley2009ros}  which provide tools for development of robotics applications. There have been several specialized robotics toolkits as well for motion planning~\cite{chitta2012moveit,ompl}, SLAM~\cite{murORB2} and manipulation~\cite{openrave}. We refer readers to Tsardoulias and Mitkas~\cite{emmanouil2017} for a comprehensive review on robotics platforms.  However, a lot of these frameworks and toolkits have still required substantial robotics expertise and highly expensive hardware. In order to make robotics more accessible and available to rest of the AI community, frameworks such as  PyRobot~\cite{pyrobot} have been introduced. But the core focus is still on robotics platforms and algorithms. We believe that with recent successes in several subfields of AI, the stage has been set to go back to grand goals of the past~\cite{MITCopyDemo}  and bring together action, perception, language and cognition into one common framework. The \pf{}  attempts to do that by developing a platform to build agents with these capabilities.

\section{Conclusion}
We introduce the \pf{}. The goal of our platform is to build agents that learn continuously from the real world interactions. The key characteristics of our architecture are: (a) modular design and heterogeneous supervision allows us to use it all: real interactions, static datasets and even heuristics; (b) multi-modality: a common platform for perception, language, and action.   We hope the \pf{} opens up new avenues of research in self-supervised learning, multi-modal learning, interactive learning, human-robot interaction and lifelong learning.


\bibliography{refs}

\begin{thebibliography}{10}
\providecommand{\url}[1]{#1}
\csname url@samestyle\endcsname
\providecommand{\newblock}{\relax}
\providecommand{\bibinfo}[2]{#2}
\providecommand{\BIBentrySTDinterwordspacing}{\spaceskip=0pt\relax}
\providecommand{\BIBentryALTinterwordstretchfactor}{4}
\providecommand{\BIBentryALTinterwordspacing}{\spaceskip=\fontdimen2\font plus
\BIBentryALTinterwordstretchfactor\fontdimen3\font minus
  \fontdimen4\font\relax}
\providecommand{\BIBforeignlanguage}[2]{{%
\expandafter\ifx\csname l@#1\endcsname\relax
\typeout{** WARNING: IEEEtranS.bst: No hyphenation pattern has been}%
\typeout{** loaded for the language `#1'. Using the pattern for}%
\typeout{** the default language instead.}%
\else
\language=\csname l@#1\endcsname
\fi
#2}}
\providecommand{\BIBdecl}{\relax}
\BIBdecl

\bibitem{tensorflow}
M.~Abadi, A.~Agarwal, P.~Barham, E.~Brevdo, Z.~Chen, C.~Citro, G.~S. Corrado,
  A.~Davis, J.~Dean, M.~Devin \emph{et~al.}, ``Tensorflow: Large-scale machine
  learning on heterogeneous distributed systems,'' \emph{arXiv preprint
  arXiv:1603.04467}, 2016.

\bibitem{pulkitpoking}
\BIBentryALTinterwordspacing
P.~Agarwal, A.~Nair, P.~Abbeel, J.~Malik, and S.~Levine, ``Learning to poke by
  poking: Experiential learning of intuitive physics,'' 2016. [Online].
  Available: \url{http://arxiv.org/abs/1606.07419}
\BIBentrySTDinterwordspacing

\bibitem{andreas2016neural}
J.~Andreas, M.~Rohrbach, T.~Darrell, and D.~Klein, ``Neural module networks,''
  in \emph{Proceedings of the IEEE conference on computer vision and pattern
  recognition}, 2016, pp. 39--48.

\bibitem{rt-track}
\BIBentryALTinterwordspacing
A.~Bewley, Z.~Ge, L.~Ott, F.~Ramos, and B.~Upcroft, ``Simple online and
  realtime tracking,'' \emph{CoRR}, vol. abs/1602.00763, 2016. [Online].
  Available: \url{http://arxiv.org/abs/1602.00763}
\BIBentrySTDinterwordspacing

\bibitem{brooks}
R.~Brooks, ``A robust layered control system for a mobile robot,'' \emph{AI
  Memo 864}, 1985.

\bibitem{SimCLR}
T.~Chen, S.~Kornblith, M.~Norouzi, and G.~Hinton, ``A simple framework for
  contrastive learning of visual representations,'' 2020.

\bibitem{NEIL}
X.~Chen, A.~Shrivastava, and A.~Gupta, ``Neil: Extracting visual knowledge from
  web data,'' in \emph{Proceedings of the IEEE international conference on
  computer vision}, 2013, pp. 1409--1416.

\bibitem{chitta2012moveit}
S.~Chitta, I.~Sucan, and S.~Cousins, ``Moveit![ros topics],'' \emph{IEEE
  Robotics \& Automation Magazine}, vol.~19, no.~1, pp. 18--19, 2012.

\bibitem{dalmia2019enforcing}
S.~Dalmia, A.~Mohamed, M.~Lewis, F.~Metze, and L.~Zettlemoyer, ``Enforcing
  encoder-decoder modularity in sequence-to-sequence models,'' \emph{arXiv
  preprint arXiv:1911.03782}, 2019.

\bibitem{devin2017learning}
C.~Devin, A.~Gupta, T.~Darrell, P.~Abbeel, and S.~Levine, ``Learning modular
  neural network policies for multi-task and multi-robot transfer,'' in
  \emph{2017 IEEE International Conference on Robotics and Automation
  (ICRA)}.\hskip 1em plus 0.5em minus 0.4em\relax IEEE, 2017, pp. 2169--2176.

\bibitem{BERT}
J.~Devlin, M.-W. Chang, K.~Lee, and K.~Toutanova, ``Bert: Pre-training of deep
  bidirectional transformers for language understanding,'' \emph{arXiv preprint
  arXiv:1810.04805}, 2018.

\bibitem{openrave}
R.~Diankov and J.~Kuffner, ``Openrave: A planning architecture for autonomous
  robotics,'' 2008.

\bibitem{Doersch2015}
\BIBentryALTinterwordspacing
C.~Doersch, A.~Gupta, and A.~A. Efros, ``Unsupervised visual representation
  learning by context prediction,'' \emph{CoRR}, vol. abs/1505.05192, 2015.
  [Online]. Available: \url{http://arxiv.org/abs/1505.05192}
\BIBentrySTDinterwordspacing

\bibitem{chelseapushing}
C.~Finn, I.~Goodfellow, and S.~Levine, ``Unsupervised learning for physical
  interaction through video prediction,'' in \emph{Advances in neural
  information processing systems}, 2016.

\bibitem{vash2020}
M.~H. Fischer, G.~Campagna, E.~Choi, and M.~S. Lam, ``Multi-modal end-user
  programming of web-based virtual assistant skills,'' 2020.

\bibitem{Gardner2017AllenNLP}
M.~Gardner, J.~Grus, M.~Neumann, O.~Tafjord, P.~Dasigi, N.~F. Liu, M.~Peters,
  M.~Schmitz, and L.~S. Zettlemoyer, ``Allennlp: A deep semantic natural
  language processing platform,'' 2017.

\bibitem{Detectron}
R.~Girshick, I.~Radosavovic, G.~Gkioxari, P.~Doll\'{a}r, and K.~He,
  ``Detectron,'' \url{https://github.com/facebookresearch/detectron}, 2018.

\bibitem{craftassist}
J.~Gray, K.~Srinet, Y.~Jernite, H.~Yu, Z.~Chen, D.~Guo, S.~Goyal, C.~L.
  Zitnick, and A.~Szlam, ``Craftassist: A framework for dialogue-enabled
  interactive agents,'' \emph{arXiv preprint arXiv:1907.08584}, 2019.

\bibitem{robotsinhome}
A.~Gupta, A.~Murali, D.~Gandhi, and L.~Pinto, ``Robot learning in homes:
  Improving generalization and reducing dataset bias,'' in \emph{Advances in
  neural information processing systems}, 2018.

\bibitem{gupta2019lvis}
A.~Gupta, P.~Dollár, and R.~Girshick, ``Lvis: A dataset for large vocabulary
  instance segmentation,'' 2019.

\bibitem{maskrcnn}
K.~{He}, G.~{Gkioxari}, P.~{Dollár}, and R.~{Girshick}, ``Mask r-cnn,'' in
  \emph{2017 IEEE International Conference on Computer Vision (ICCV)}, 2017,
  pp. 2980--2988.

\bibitem{MoCo}
K.~He, H.~Fan, Y.~Wu, S.~Xie, and R.~Girshick, ``Momentum contrast for
  unsupervised visual representation learning,'' 2020.

\bibitem{MITCopyDemo}
B.~Horn, B.~Klaus, and P.~Horn, ``Section 15.7: The copy demonstration,'' in
  \emph{Robot vision}, 1986.

\bibitem{modelzoo}
J.~Y. Koh, ``Model zoo,'' \url{https://modelzoo.co/}, 2020.

\bibitem{coco}
\BIBentryALTinterwordspacing
T.~Lin, M.~Maire, S.~J. Belongie, L.~D. Bourdev, R.~B. Girshick, J.~Hays,
  P.~Perona, D.~Ramanan, P.~Doll{\'{a}}r, and C.~L. Zitnick, ``Microsoft
  {COCO:} common objects in context,'' \emph{CoRR}, vol. abs/1405.0312, 2014.
  [Online]. Available: \url{http://arxiv.org/abs/1405.0312}
\BIBentrySTDinterwordspacing

\bibitem{thinsoftware}
A.~Makarenko, A.~Brooks, and T.~Kaupp, ``On the benefits of making robotic
  software frameworks thin,'' in \emph{IEEE/RSJ International Conference on
  Intelligent Robots and Systems (IROS)}, 2007.

\bibitem{habitat19iccv}
{Manolis Savva*}, {Abhishek Kadian*}, {Oleksandr Maksymets*}, Y.~Zhao,
  E.~Wijmans, B.~Jain, J.~Straub, J.~Liu, V.~Koltun, J.~Malik, D.~Parikh, and
  D.~Batra, ``Habitat: {A} {P}latform for {E}mbodied {AI} {R}esearch,'' in
  \emph{Proceedings of the IEEE/CVF International Conference on Computer Vision
  (ICCV)}, 2019.

\bibitem{word2vec}
T.~Mikolov, I.~Sutskever, K.~Chen, G.~S. Corrado, and J.~Dean, ``Distributed
  representations of words and phrases and their compositionality,'' in
  \emph{Advances in neural information processing systems}, 2013, pp.
  3111--3119.

\bibitem{NELL}
T.~Mitchell, W.~Cohen, E.~Hruschka, P.~Talukdar, B.~Yang, J.~Betteridge,
  A.~Carlson, B.~Dalvi, M.~Gardner, B.~Kisiel \emph{et~al.}, ``Never-ending
  learning,'' \emph{Communications of the ACM}, vol.~61, no.~5, pp. 103--115,
  2018.

\bibitem{pytorchmodules}
P.~Modules, ``Pytorch modules,''
  \url{https://pytorch.org/docs/stable/generated/torch.nn.Module.html}, 2020.

\bibitem{tfmodules}
T.~Modules, ``Tensorflow modules,''
  \url{https://www.tensorflow.org/guide/intro_to_modules}, 2020.

\bibitem{murORB2}
R.~Mur-Artal and J.~D. Tard\'os, ``{ORB-SLAM2}: an open-source {SLAM} system
  for monocular, stereo and {RGB-D} cameras,'' \emph{IEEE Transactions on
  Robotics}, vol.~33, no.~5, pp. 1255--1262, 2017.

\bibitem{pyrobot}
A.~Murali, T.~Chen, K.~V. Alwala, D.~Gandhi, L.~Pinto, S.~Gupta, and A.~Gupta,
  ``Pyrobot: An open-source robotics framework for research and benchmarking,''
  2019.

\bibitem{pytorch}
A.~Paszke, S.~Gross, F.~Massa, A.~Lerer, J.~Bradbury, G.~Chanan, T.~Killeen,
  Z.~Lin, N.~Gimelshein, L.~Antiga, A.~Desmaison, A.~Kopf, E.~Yang, Z.~DeVito,
  M.~Raison, A.~Tejani, S.~Chilamkurthy, B.~Steiner, L.~Fang, J.~Bai, and
  S.~Chintala, ``Pytorch: An imperative style, high-performance deep learning
  library,'' in \emph{Advances in Neural Information Processing Systems 32},
  H.~Wallach, H.~Larochelle, A.~Beygelzimer, F.~d\textquotesingle
  Alch\'{e}-Buc, E.~Fox, and R.~Garnett, Eds.\hskip 1em plus 0.5em minus
  0.4em\relax Curran Associates, Inc., 2019, pp. 8024--8035.

\bibitem{Pinto2016Grasping}
L.~Pinto and A.~Gupta, ``Supersizing self-supervision: Learning to grasp from
  50k tries and 700 robot hours,'' 2015.

\bibitem{quigley2009ros}
M.~Quigley, K.~Conley, B.~Gerkey, J.~Faust, T.~Foote, J.~Leibs, R.~Wheeler, and
  A.~Y. Ng, ``{ROS}: an open-source robot operating system,'' in \emph{ICRA
  workshop on open source software}, vol.~3, no. 3.2.\hskip 1em plus 0.5em
  minus 0.4em\relax Kobe, Japan, 2009, p.~5.

\bibitem{blenderbot}
S.~Roller, E.~Dinan, N.~Goyal, D.~Ju, M.~Williamson, Y.~Liu, J.~Xu, M.~Ott,
  K.~Shuster, E.~M. Smith \emph{et~al.}, ``Recipes for building an open-domain
  chatbot,'' \emph{arXiv preprint arXiv:2004.13637}, 2020.

\bibitem{facenet}
\BIBentryALTinterwordspacing
F.~Schroff, D.~Kalenichenko, and J.~Philbin, ``Facenet: {A} unified embedding
  for face recognition and clustering,'' \emph{CoRR}, vol. abs/1503.03832,
  2015. [Online]. Available: \url{http://arxiv.org/abs/1503.03832}
\BIBentrySTDinterwordspacing

\bibitem{mmf}
A.~Singh, V.~Goswami, V.~Natarajan, Y.~Jiang, X.~Chen, M.~Shah, M.~Rohrbach,
  D.~Batra, and D.~Parikh, ``Mmf: A multimodal framework for vision and
  language research,'' \url{https://github.com/facebookresearch/mmf}, 2020.

\bibitem{caip2020}
\BIBentryALTinterwordspacing
K.~Srinet, Y.~Jernite, J.~Gray, and A.~Szlam, ``{C}raft{A}ssist instruction
  parsing: Semantic parsing for a voxel-world assistant,'' in \emph{Proceedings
  of the 58th Annual Meeting of the Association for Computational
  Linguistics}.\hskip 1em plus 0.5em minus 0.4em\relax Online: Association for
  Computational Linguistics, Jul. 2020, pp. 4693--4714. [Online]. Available:
  \url{https://www.aclweb.org/anthology/2020.acl-main.427}
\BIBentrySTDinterwordspacing

\bibitem{ompl}
I.~Sucan, M.~Moll, and L.~Kavraki, ``The open motion planning library,''
  \emph{IEEE Robotics \& Automation Magazine}, vol.~19, no.~4, 2012.

\bibitem{emmanouil2017}
E.~Tsardoulias and P.~Mitkas, ``Robotic frameworks, architectures and
  middleware comparison,'' \emph{arXiv preprint arXiv:1711.06842}, 2017.

\bibitem{pytorchtutorial}
P.~Tutorial, ``Pytorch tutorial,'' \url{https://pytorch.org/tutorials/}, 2020.

\bibitem{tftutorial}
T.~Tutorial, ``Tensorflow tutorial,''
  \url{https://www.tensorflow.org/tutorials}, 2020.

\bibitem{wang2017naturalizing}
S.~I. Wang, S.~Ginn, P.~Liang, and C.~D. Manning, ``Naturalizing a programming
  language via interactive learning,'' \emph{arXiv preprint arXiv:1704.06956},
  2017.

\bibitem{Wang2015}
X.~Wang and A.~Gupta, ``Unsupervised learning of visual representations using
  videos,'' in \emph{Proceedings of the IEEE international conference on
  computer vision}, 2015, pp. 2794--2802.

\end{thebibliography}
\bibliographystyle{IEEEtranS}


\end{document}